\documentclass{article}

\usepackage{PRIMEarxiv}

\usepackage[utf8]{inputenc} % allow utf-8 input
\usepackage[T1]{fontenc}    % use 8-bit T1 fonts
\usepackage{hyperref}       % hyperlinks
\usepackage{url}            % simple URL typesetting
\usepackage{booktabs}       % professional-quality tables
\usepackage{amsfonts}       % blackboard math symbols
\usepackage{nicefrac}       % compact symbols for 1/2, etc.
\usepackage{microtype}      % microtypography
\usepackage{lipsum}
\usepackage{fancyhdr}       % header
\usepackage{graphicx}       % graphics
\graphicspath{{media/}}     % organize your images and other figures under media/ folder
\usepackage{pgfplots}
\usepackage{caption}     % Load first
\usepackage{subcaption}
\usepackage{tikz}
\usepgfplotslibrary{groupplots}
\usepackage{pdflscape}
\usepackage{adjustbox}
\usepackage{booktabs}
\usepackage{amsmath}
\usepackage{color,soul}
\usepackage{subcaption}
\usepackage{algorithm}
\usepackage{algpseudocode}
\usepackage{tabularx}
\title{Tabular Diffusion based Actionable Counterfactual Explanations for Network Intrusion Detection
}

\author{
  Vinura Galwaduge, Jagath Samarabandu \\
  Dept. of Electrical and Computer Engineering,
  University of Western Ontario\\
  London\\
  \texttt{\{vgalwadu, jagath\}@uwo.ca} \\
}

\begin{document}
\maketitle

\begin{abstract}
Modern network intrusion detection systems (NIDS) frequently utilize the predictive power of complex deep learning models. However, the "black-box" nature of such deep learning methods adds a layer of opaqueness that hinders the proper understanding of detection decisions, trust in the decisions and prevent timely countermeasures against such attacks. Explainable AI (XAI) methods provide a solution to this problem by providing insights into the causes of the predictions. The majority of the existing XAI methods provide explanations which are not convenient to convert into actionable countermeasures. In this work, we propose a novel diffusion-based counterfactual explanation framework that can provide actionable explanations for network intrusion attacks. We evaluated our proposed algorithm against several other publicly available counterfactual explanation algorithms on 3 modern network intrusion datasets. To the best of our knowledge, this work also presents the first comparative analysis of existing counterfactual explanation algorithms within the context of network intrusion detection systems. Our proposed method provide minimal, diverse counterfactual explanations out of the tested counterfactual explanation algorithms in a more efficient manner by reducing the time to generate explanations. We also demonstrate how counterfactual explanations can provide actionable explanations by summarizing them to create a set of global rules. These rules are actionable not only at instance level but also at the global level for intrusion attacks. These global counterfactual rules show the ability to effectively filter out incoming attack queries which is crucial for efficient intrusion detection and defense mechanisms.
\end{abstract}

% keywords can be removed

\section{Introduction}
Deep learning-based network intrusion detection systems (NIDS) plays a crucial role in the modern cybersecurity landscape, mainly due to the generalizability and scalability of deep learning models for high-dimensional network telemetry data~\cite{gamage_deep_2020}. Nonetheless, the opaque nature of these deep learning models presents a significant challenge for human analysts at Security Operations Centers (SoCs) to determine the root causes of threats and formulate appropriate countermeasures against intrusion attempts.  Explainable AI (XAI) techniques are generally proposed as an effective solution for this lack of opaqueness issue across numerous application domains. In the context of NIDS, feature attribution methods such as Shapley Additive Explanations (SHAP), and Local Interpretable Model-Agnostic Explanations (LIME) are frequently used to explain decisions made by NIDS~\cite{houda_why_2022}. Feature attribution methods usually provide a ranking of features based on a certain type of feature score. The features score is derived based on the model's outputs (i.e.-predictions) for a data point or a group of data points. For example, Shapley values provide feature scores as a measurement of the contribution of each feature to the prediction of a black-box model~\cite{lundberg_unified_2017}. LIME also calculates a feature importance score, based on a simple surrogate model such as a logistic regression model trained on synthetically sampled instances around a data point that needs to be explained. However, a major shortcoming of such feature attribution methods is that they do not provide any actionable recourse for the human analysts to work on. Feature attribution methods primarily help human analysts to identify the important features. Analysts would then use these explanations to identify the type of the threat and design appropriate defense mechanisms. However, this entire process can consume a substantial amount of time which is critical to act against the threat and ensure normal operations without incurring losses. On the other hand, emerging paradigms such as "Zero-touch Networks" demand the ability to automate the process of defense against intrusion attacks, without having to rely on the expertise of human analysts~\cite{benzaidAIDriven2020}.\\
Counterfactual explanations bear the advantages of feature attribution methods as well as the potential to provide actionable countermeasures within the explanations. The traditional definition of counterfactual explanation describes that a counterfactual explanation should produce an alternate data point that has a different (or a preferred) prediction from the black box model, with minimal changes to the original data point that needs to be explained~\cite{wachter_counterfactual_2018}. In the context of NIDS that has classified a certain network activity as an intrusion, a counterfactual explanation might involve generating a new data instance that the black-box decision model classifies as a normal (non-intrusion) activity. This counterfactual instance highlights the minimal set of feature modifications required to alter the original prediction, thereby revealing both the influential features contributing to the model’s decision and the specific value adjustments needed. These value adjustments can then be converted to actions against the attack.\\
A diverse range of counterfactual explanation methods has been proposed in the literature for algorithmic recourse, mainly concerning financial domain (e.g.- loan application scenarios) or recidivism as the real-world applications~\cite{guidotti_counterfactual_2022, verma_counterfactual_2021, stepin_survey_2021}. These multitude of counterfactual algorithms also provide different approaches to generate counterfactual explanations carrying different properties. For instance, sparsity (minimal changes) of counterfactual explanations have been generally preferred while certain types of algorithms prefer plausibility (explanations following the underlying distribution of the original data) over sparsity~\cite{pawelczyk_counterfactual_2020}. This difference of qualities can be observed across domains. For example, finance or recidivism domains would prioritize fairness of counterfactuals more often while counterfactual explanations for debugging applications might not do so~\cite{gupta_equalizing_2019}. Hence, based on this observation, we focus more on efficient, generative counterfactual generation approaches such as 'in-training' methods as suitable for NIDS~\cite{guyomard_vcnet_2022}.
Overall, our main contributions in this work are listed as below.
\begin{enumerate}
    \item We propose a novel diffusion based algorithm to generate counterfactual explanations for network intrusion detection applications. We provide empirical evidences of the utility of the proposed method compared to several existing, publicly available counterfactual algorithms.
    \item To the best of our knowledge, this quantitative comparison of existing counterfactual methods provide the first of an analysis of its kind in the literature, for NIDS domain.
    \item Finally, we propose global counterfactual rules to summarize the counterfactual explanations generated by the proposed method. These rules are derived using a simple yet an effective technique based on decision-trees. These rules summarize the important features for a cohort of attack data points, and the associated value bounds that separate them from benign data. Hence, these rules can be used in actionable defense measures against network intrusion attacks. 
\end{enumerate}

In this study, we constrain the downstream task to a supervised binary classification setting, focusing on generating counterfactual explanations for data instances that are classified as attacks.

section{Related Work}

Counterfactual explanations have been recently gaining traction among practitioners mainly due to their ability to provide actionable explanations. The inherent qualities of counterfactual explanations can vary depending on the algorithmic approach employed for their generation~\cite{guidotti_counterfactual_2022}. For example, a list of the types of model-agnostic algorithms are presented where each type of approach offer different qualities of counterfactual explanations~\cite{chou_counterfactuals_2022}. A few recent attempts show how certain algorithms impact the quality of the generated counterfactual explanations~\cite{leemann_towards_2024, ferry_sok_2023}. Owing to the impact of AI regulatory frameworks~\cite{noauthor_general_nodate} and the early literature~\cite{wachter_counterfactual_2018}, most of the research on counterfactual explanations revolve around domains such as finance and criminal justice~\cite{wachter_counterfactual_2018}. Nevertheless, counterfactual explanations have also been utilized in other domains/data-modalities such as time-series data, image image and use-cases such as model debugging, and knowledge extraction.\\

The Network Intrusion Detection (NID) literature has recently focused more on providing explainability along with intrusion detection. Among the plethora of proposals for integrating XAI into NID, feature importance methods have been of particular interest. A combination of three feature attribution methods (SHAP, LIME, and RuleFit) are proposed for NID in Internet of Things (IoT)~\cite{houda_why_2022}. These feature importance methods provide the most critical features that affected each decision either locally, or globally. LIME and SHAP are again used in IoT NID, and the generated explanations show that the proposed model is aware of the protocol specific features for the classification of each attack~\cite{krishnan_explainable_nodate}. In addition to popularly used LIME and SHAP methods, Partial Dependence Plots and Individual Conditional Expectation plots have also been utilized~\cite{keshk_explainable_2023}. A comprehensive survey regarding the utility of XAI for NID in IoT domain is presented by Moustafa et al.~\cite{moustafa_explainable_2023} and a number of similar feature based explanation frameworks can be found in the literature~\cite{barnard_robust_2022, kalakoti_evaluating_2025, naif_alatawi_enhancing_2025}.\\

Although feature attribution methods have been frequently used to produce explanations, counterfactual explanations have been only sparingly considered in the NID domain. As the earliest related attempt to use counterfactual explanations, a method for generating adversarial examples is presented with the purpose of providing explanations for misclassified network activities~\cite{marino_adversarial_2018}. This closely aligns with the concept of counterfactual examples; however, the adversarial example approach lacks the provision of meaningful or actionable alternatives that human analysts can interpret or implement effectively. Moreover, their approach is limited as an analysis tool for known misclassified samples. Counterfactual inference (not explanations) have been used to identify different Distributed Denial of Service (DDoS) attack types~\cite{zeng_intrusion_2022}. There, counterfactual inference is used to identify the anomalous features that could have been caused by a specific DDoS attack type. In one study, different types of feature attribution methods and counterfactual explanations are briefly compared against each other for IoT NID~\cite{gyawali_leveraging_2024}. However, they dismiss the diversity aspect of counterfactual explanations and limit the analysis counterfactuals to one type of algorithm. However, we show that the diversity aspect will be helpful to provide multiple different explanations for an incoming attack data point. They mainly focus on using calculating SHAP feature attribution values calculated using nearest counterfactuals (CF-SHAP) which can be computationally inefficient as well. More recently, a novel approach for generating counterfactual explanations based on an energy minimization concept have been proposed with a focus on NID in IoT~\cite{evangelatos_exploring_2025}. The method involves finding minimally perturbed instances around a local neighborhood of the classifier function which is approximated by a Taylor expansion of the local decision boundary. They even carry out a robustness analysis of the proposed method. However, their work seems to lack a comparative analysis involving commonly used counterfactual evaluation metrics and methods. In contrast, this work provides a comparison between several, publicly available counterfactual methods in terms of multiple evaluation metrics.

\section{Counterfactual explanations for Network Intrusion Detection}

As observed in the existing literature, counterfactual explanations can possess different properties or qualities such as sparsity, validity and plausibility. Hence, it is imperative to identify different qualities expected by the counterfactual explanations for network intrusion detection applications. We propose a list of qualities that would be desired for network intrusion detection applications along with initial justifications to them as below.
\begin{enumerate}
    \item Efficiency - Modern cybersecurity operations centers are fast-paced and handle dynamic threat landscapes. Such environment requires efficient explanation generation in order to quickly identify threats and design countermeasures.
    \item Validity - The generated explanations are required to be valid (reside in the intended target class), which would otherwise reduce the utility of generated explanations for defense measures.
    \item Diversity - Intrusion attacks generally comprise of multiple attack queries (e.g.- malformed HTTP requests in a Denial of Service (DoS) attack). Generating multiple explanations per query can offer different perspectives into attacks than one explanation per query. Later, we use this quality to aggregate and produce global rules using diverse explanations.
    \item Sparsity and plausibility - Sparsity ensures minimal changes to the original and plausibility ensures that the generated counterfactual explanations are realistic. These two qualities can provide human analysts with explanations that are easier to analyze and utilize for countermeasures against threats. 
\end{enumerate}

In the following sections, we identify existing counterfactual algorithms that fulfills these qualities.

\subsection{Counterfactual algorithms for intrusion detection}

Based on the recent literature on counterfactual explanation methods, we identify that generative counterfactual methods - specifically the class of counterfactual explanations named 'in-training' methods as potential candidates for network intrusion detection domain. This is due to the fact that these methods can efficiently sample counterfactual explanations from learned distributions, and produce plausible counterfactual explanations while maintaining good validity~\cite{guyomard_vcnet_2022,guo_counternet_2023}. Simply described, in-training methods refer to training the counterfactual generation algorithm with the knowledge of the decision boundary of the classifier model . VCNet~\cite{guyomard_vcnet_2022} is one of the two in-training counterfactual methods available in the literature, where a Conditional Variational Auto Encoder (CVAE) is trained in-tandem with a classifier model. The  other approach utilizes a simple feed-forward neural network instead of a CVAE~\cite{guo_counternet_2023}. However, This would not be scalable for complex datasets used in domains such as NIDS.\\
More recently, diffusion models have been incorporated in tabular (structured) data domains due to their success in image generation domain~\cite{dhariwal_diffusion_2021}. Structured Counterfactual Diffuser (SCD) is presented as a counterfactual explanation method for tabular data and has shown promising results in efficiently generating diverse counterfactual explanations for simple tabular datasets~\cite{madaan_navigating_2023}. Inspired by this, we propose an improved version of a diffusion based counterfactual explanation method for network intrusion detection. This proposed improved version treats the heterogeneous features in the tabular, network intrusion datasets without transforming them into a continuous latent space~\cite{madaan_navigating_2023}. The proposed model is then further improved with the help of 'Progressive Distillation'~\cite{salimans_progressive_2022}. This distillation step primarily reduces the time to generate counterfactual explanations by reducing the number of diffusion steps required. In addition, this distillation step yields improvements in sparsity of the explanations as well. In this work, we reduce the number of diffusion steps by tenfold which is explained further in the succeeding subsection (section~\ref{sub:diff}). 

\subsection{Guided diffusion based counterfactual explanations for network intrusion detection}

SCD utilizes diffusion guidance to generate counterfactual explanations. This approach is somewhat similar to in-training methods where the knowledge of the decision boundary is used to train the counterfactual generation method. In contrast to SCD's utilization of a shared embedding layer to encode both continuous and categorical types of features, we treat numerical and categorical features separately in our proposed model. We provide an overview of the theoretical background for the VCNet model and the guided diffusion model proposed in this work in the following section.

\subsubsection{VCNet model}
VCNet model utilize a Conditional Variational Auto-Encoder (CVAE) to learn the conditional data distributions during the training phase. The classifier (black-box model) is also trained alongside the CVAE and the classifier's prediction outputs are fed to the CVAE as a conditional signal. This feedback path is used for generating counterfactual explanations during the inference stage. The counterfactual explanation ($x_{CF}$) for an input data instance $x$ is generated using the trained CVAE with the help of the target label information. This is mathematically represented by Eq~\ref{eq:vcnet}, where $\Theta$ denotes the trained CVAE decoder, $y'$ the target label, $y$ the original label, and the latent representations of the explanation and the original data instance by $z'$ and $z$.

\begin{equation}
    \label{eq:vcnet}
    x_{CF} = \Theta(z'|y'), z' \in \mathcal{N}(z|x,y)\\
\end{equation}

In our experiments, we replace the simple isotropic Gaussian prior of the CVAE with a mixture of Gaussian prior. This is due to our initial findings that showed a mixture of Gaussian can improve the validity of the produced counterfactual explanations without hindering other performance metrics.

\subsubsection{Guided diffusion models}
\label{sub:diff}

Denoising diffusion probabilistic models (DDPM) are built on the concept of diffusion - gradually (stepwise) destroying the structure of an input data distribution until it takes the form of a known prior distribution, such as the isotropic Gaussian~\cite{ho_denoising_2020}. This process is called the forward diffusion process. During sampling, a reverse diffusion process, that is, reversing the initial action performed by the forward diffusion process, is applied to samples taken from the known prior in order to generate new samples from the original data distribution. In the context of tabular datasets, we propose using separate diffusion processes for numerical and categorical features, inspired by the TabDDPM model~\cite{kotelnikov_tabddpm_2022}. These separate diffusion processes ensure that numerical features are diffused using Gaussian noise while the categorical features are diffused using categorical noise. For a tabular dataset $D$ with $N_{num}$ numerical features and $N_{cat}$ categorical features, the forward diffusion process for numerical features is formulated as,

\begin{equation}
\label{eq1}
x_{t,num} = \mathcal(\sqrt{\bar\alpha_t}x_{0,num},(1-\bar\alpha_t)\epsilon), \forall x_{0,num} \in [0,N_{num}]
\end{equation}
where,~$\alpha =\prod_{i=1}^{t}\beta_i$, and $\epsilon \in \mathcal{N}(0,1)$. $\beta_i$ refers to forward process variance and $i\in [0,T]$ denotes the diffusion steps.\\
Forward diffusion process for categorical features is represented as, 

\begin{equation}
\label{eq2}
    x_{t,cat_i} = \mathcal{C}(\bar\alpha_{0,cat_i}+(1-\bar\alpha)/K)
\end{equation}
where, $K$ is the categories of feature $i$ and $i \in [0,N_{cat}]$.
During sampling, counterfactual explanations are generated using the guidance from a classifier model, which was trained using the outputs of the trained diffusion model. The classifier guidance is in the form of a gradient value which is calculated as the gradient of a counterfactual explanation loss function (Eq.~\ref{eq3}) with respect to the intermediate outputs of the diffusion model. We modify the counterfactual loss function proposed in SCD, to a simpler version. 

\begin{equation}
\label{eq3}
    \mathcal{L} = [y_{CF}log(f(x_t))+(1-y_{CF})(1-log((f(x_t)))]+d(x_t,x_{orig})
\end{equation}

where $T$ denotes the diffusion steps, $t \in [0,T]$. The term inside the square brackets is the binary cross entropy loss, whereas $y_{CF}$, and $x_{orig}$ denote the target prediction of the counterfactual explanation and the original query, respectively. A distance measuring function is denoted by $d$ that penalizes explanations generated far away from the original query. L1-norm is used for $d$ in our proposed method. Counterfactual generation process is carried out by first creating a noisy version of the original data point ($x_{0,noise}$) using Eq~\ref{eq1} and Eq~\ref{eq2}. Then, the reverse diffusion process is carried out using classifier guidance (Eq~\ref{eq4}), until a counterfactual explanation $x_{CF_t}, t=0$ is reached. The classifier guidance in Eq~\ref{eq4} simply guides the intermediate representations ($x_{t+1}$) toward the counterfactual explanation ($x_{CF_t}$), using the scaled gradient of the loss function(Eq~\ref{eq3}) w.r.t to $x_{t+1}$. The scalar value ($\alpha$) can be selected arbitrarily such that $\alpha \in \mathbb{R}$.  
The proposed diffusion methods and SCD inherently provides the capability of obtaining multiple counterfactual explanations during the generation process.These are also known as 'diverse' counterfactual explanations. These diverse counterfactuals are then utilized to generate global counterfactual rules, as shown later in results (section~\ref{sec:res}). This ability of diffusion models stands-out as they can efficiently generate multiple explanations at a time.

\begin{equation}
\label{eq4}
    x_{CF_{t+1}} \leftarrow x_{t+1} - \alpha\nabla_{x_{t+1}}(\mathcal{L})
\end{equation}

Afterward, we further refine the diffusion model through 'Progressive Distillation' which reduces the number of diffusion steps required to generate counterfactual explanations~\cite{salimans_progressive_2022}. Using distillation, we were able to reduce the number of steps by tenfold (2500 to 250) as shown in the results (section~\ref{sec:res}). Although further reduction is possible, in this work we adopt a tenfold reduction due to the trade-off observed between computational time and plausibility of explanations. Theoretical background regarding the distillation method are presented in the Appendix (App. ~\ref{app1}). 

\section{Methodology}

We consider the downstream classification task as a binary classification task (benign vs attack), and generate counterfactual explanations for a randomly selected pool of attack queries from the test dataset, during the explanation generation stage. An average of the metrics across 5 different random runs is obtained, where each run is controlled by a distinct random seed. Quantitative evaluations are carried out for increasing amounts of the random pool and the max size of a random pool is limited to 4000 data points due to resource and time constraints. All the experiments were carried out on a system containing a single NVIDIA-RTX-2080 GPU with 12GB memory.

\subsection{Datasets}

Three popular used network intrusion detection datasets are selected; UNSW-NB15~\cite{moustafa_unsw-nb15_2015}, CIC-IDS-2017~\cite{sharafaldin_toward_2018}, and CIC-DDoS-2019~\cite{sharafaldin_developing_2019}. Each dataset comprises of benign and attack queries, derived based on network flow data, and multiple categories of attack queries. The datasets are preprocessed by removing highly-correlated features. This step ensures that the performance of the black-box classifier is improved, and the generated explanations are more compact. A 'K-Bins Discretizer' is used for scaling the numerical features for the SCD model (following the original work), and 'Quantile Transformer' for the rest of the models. This was selected following the approach used in TabDDPM model. One-Hot Encoder is used to preprocess the categorical features. An overview of the dataset statistics after preprocessing are provided in Table~\ref{tab:data}. We use the dataset files of these intrusion detection datasets publicly available on Kaggle\footnote{https://www.kaggle.com/dhoogla/datasets}.

% We provide additional details about the datasets in Appendix~\ref{appB}.
\begin{table}[]
    \centering
    \begin{tabular}{c|c|c}
        \hline
         Dataset&Features&No. of instances  \\
         \hline
         UNSW-NB15&34 (31 num., 3 cat.)&257 499\\
         CICDDoS-2019& 32 (26 num., 6 cat.)&190 383\\
         CICIDS-2017&49 (49 num.)&2 756 124\\
         \hline
    \end{tabular}
    \caption{Summary of the three datasets used in experiments.}
    \label{tab:data}
\end{table}

\subsection{Baseline counterfactual methods}

We use publicly available implementations of 
Wachter~\cite{wachter_counterfactual_2018}, DiCE~\cite{mothilal_explaining_2020}, FACE~\cite{poyiadzi_face_2020}, and CCHVAE~\cite{pawelczyk_learning_2020}. These implementations were adopted from the counterfactual explanations benchmarking tool 'CARLA'~\cite{pawelczyk_carla_2021}. VCNet and SCD models were implemented on our own as there were no publicly available implementations. A feed-forward neural network was selected as the black-box classifier model for all the experiments. The same architecture was used for the black-box classifier models across all the counterfactual generation methods. For VCNet we stick with a slightly different architecture (i.e.- different hidden layers and layer sizes) based on our initial experimental runs. Additional details about the black-box classifiers are provided in Appendix~\ref{app2}. Our code will be made publicly available on Github. 

\subsection{Evaluation metrics}

We pick the following evaluation metrics to evaluate the performance of generated counterfactual explanations by each method; k-validity, 1-validity, sparsity, Local Outlier Factor (LOF), and time to generate. These evaluation metrics were selected based on the prior work in counterfactual recourse domain~\cite{moreira_benchmarking_2024, pawelczyk_carla_2021, bayrak_evaluation_2024} and also considering the requirements such as efficiency for the NIDS domain. k-validity measures the average number of valid counterfactual explanations generated from a pool of generated explanations (denoted by k (e.g.- k=10)) for an attack query. This metric then is only applicable to counterfactual methods that are capable of generating multiple (diverse) counterfactual explanations (further elaborated in results (section~\ref{sec:res})). 1-validity measures the ability of the counterfactual algorithm to generate at least one valid explanation for each of the original queries. This is also calculated as an average across the dataset. Sparsity measures the average number of feature changes (L0 norm), and LOF measures the plausibility of the generated explanations. Time to generate measures either the number of seconds taken to generate one explanation or a group of explanations (if it is allowed by the counterfactual method).

\section{Results}
\label{sec:res}

Table~\ref{tab:unsw1} shows the results obtained for the UNSW-NB15 dataset. The Acc./F1 column depicts the performance of the black-box classifier model for each counterfactual method. We omit the results from FACE counterfactual generation method for the other two datasets as well as larger attack pools, as it took a long time (\>5 min. per explanation) to generate counterfactual explanations for an attack pool of 1000 instances from the UNSW-NB15 dataset. k-validity values are omitted for FACE and VCNet as they are unable to generate multiple (diverse) counterfactual explanations by design. CCHVAE is also limited to generating single explanation for a query. Although the authors of CCHVAE claim that it can be extended to generate multiple counterfactual explanations~\cite{pawelczyk_learning_2020} we were unable to test it since that was not implemented in CARLA. Similarly, Tables~\ref{tab:ddos1} and \ref{tab:cic1} show results for CIC-DDoS-2019 and CIC-IDS-2017 datasets. The results from CCHVAE for the last two datasets were also omitted as it failed to return counterfactual explanations for the attack queries within a reasonable amount of time. The proposed methods (TabDiff and TabDiff-distill.) consistently provide counterfactual explanations with high validity across all the datasets. VCNet model provides the best efficiency (in terms of the time to generate explanations). The distilled version of the diffusion model (TabDiff-distill.) provides the second best efficiency across the three datasets. However, VCNet lags in terms of sparsity as observed for the UNSW-NB15 dataset, in addition to being limited to a single explanation per query. This is clearly shown in figure~\ref{scatter}(a) where the TabDiff-distill. shows better sparsity, and in tables~\ref{tab:unsw1},\ref{tab:ddos1} and \ref{tab:cic1} where the VCNet model does not carry k-validity values. Moreover, the proposed methods produce plausible explanations as reflected by the comparatively low log-LOF values. The scatter plots in figure~\ref{scatter} illustrates how the performance of the counterfactual methods from tables~\ref{tab:unsw1},\ref{tab:ddos1} and \ref{tab:cic1} behave when sparsity, efficiency and validity metrics are combined with each other. 

\begin{table}[ht]
    \centering
    \begin{tabularx}{\textwidth}{X|X|X|X|X|X|X}
        \hline
        Method&Sparsity$\downarrow$&k-validity$\uparrow$&validity$\uparrow$&log-LOF$\downarrow$&time$\downarrow$&Acc/F1\\
        \hline
        Wachter&6.51\tiny$\pm$0.18&-&0\tiny$\pm$0&1.25\tiny$\pm$4.84&0.04\tiny$\pm$0.0003&85.94/87.79\\
         DiCE&33.2\tiny$\pm$0.01&8.62\tiny$\pm$0.03&0.99\tiny$\pm$0.003&6.82\tiny$\pm$0.14&2.14\tiny$\pm$0.07&89.10/90.44 \\
         FACE&1\tiny$\pm$0&-&0\tiny$\pm$0&1.21\tiny$\pm$0.01&395.75\tiny$\pm$1&87.65/89.02\\
         VCNet&33.96\tiny$\pm$0.0004&-&0.97\tiny$\pm$0.004&1.59\tiny$\pm$0.24&4.35e-6\tiny$\pm$2e-7&85.83/85.30\\
         CCHVAE&20.11\tiny$\pm$0.28&-&0\tiny $\pm$0&1.34\tiny$\pm$ 0.13&24.44\tiny $\pm$0.68&88.32/90.55\\
         SCD&22.63\tiny$\pm$0.05&2.13\tiny$\pm$0.13&0.8\tiny$\pm$0.02&0.26\tiny$\pm$0.02&3.01\tiny$\pm$0.45&87.65/89.02\\
         \textbf{TabDiff}&19.5\tiny$\pm$0.1&7.44\tiny$\pm$0.04&1\tiny$\pm$0.02&0.12\tiny$\pm$0.29&6.22\tiny$\pm$0.18&87.65/89.02\\
         \textbf{TabDiff-distill.}&17.92\tiny$\pm$0.22&5.03\tiny$\pm$0.02&1\tiny$\pm$0&0.68\tiny$\pm$0.14&0.92\tiny$\pm$0.02&87.65/89.02\\
         \hline
    \end{tabularx}
    \caption{Results for UNSW-NB15 dataset, with 1000 attack queries. The arrow directions alongside each metric denotes the preferred direction (low/high) for the metric to behave in order to get better explanations. Each evaluated metric contains its standard deviation across the 5 multiple runs ($\pm\sigma$, in tiny font) alongside its mean value. The proposed methods provide plausible counterfactuals with high validity, efficiency and sparsity.}
    \label{tab:unsw1}
\end{table}

\begin{table}[ht]
    \centering
    \begin{tabularx}{\textwidth}{X|X|X|X|X|X|X}
        \hline
        Method&Sparsity$\downarrow$&k-validity$\uparrow$&validity$\uparrow$&log-LOF$\downarrow$&time$\downarrow$&Acc/F1\\
        \hline
        Wachter&6.25\tiny$\pm$0.03&-&0\tiny$\pm$0.003&0.72\tiny$\pm$1.26&7.8e-3\tiny$\pm$0.0001&96.43/92.92\\
         DiCE&3.8\tiny$\pm$0.02&0.04\tiny $\pm$0.01&0.04\tiny$\pm$ 0.01&0.74\tiny$\pm$0.11&2.8\tiny $\pm$0.29&99.15/96.67\\
         FACE&-&-&-&-&-&-\\
         VCNet&20.24\tiny$\pm$0.02&-&0.97\tiny$\pm$0.008&1.05\tiny$\pm$0.27&4.58e-6\tiny$\pm$9.6e-7&99.13/98.24\\
         CCHVAE&-&-&-&-&-&-\\
         SCD&20.95\tiny$\pm$0.2&8.78\tiny$\pm$0.58&1\tiny$\pm$0&0.38\tiny$\pm$0.12&8.06\tiny$\pm$0.53&99.06/98.11\\
         \textbf{TabDiff}&22.59\tiny$\pm$0.01&8.92\tiny$\pm$0.001&1\tiny $\pm$ 0&0.12\tiny$\pm$0.001&15.55\tiny$\pm$0.76&99.14/98.25\\
         \textbf{TabDiff-distill.}&16.28\tiny$\pm$0.03&3.87\tiny$\pm$0.02&0.99\tiny$\pm$0.001&0.15\tiny$\pm$0.1&1.65\tiny$\pm$0.07&99.1/96.54\\
         \hline
    \end{tabularx}
    \caption{Results for CIC-DDoS-2019 dataset, with 1000 attack queries. The proposed methods provide plausible counterfactuals with high validity, efficiency and sparsity. While VCNet shows comparably good performance by providing sparse and valid explanations, it fails to do so across all the datasets consistently. }
    \label{tab:ddos1}
\end{table}

\begin{table}[ht]
    \centering
    \begin{tabularx}{\textwidth}{X|X|X|X|X|X|X}
        \hline
        Method&Sparsity$\downarrow$&k-validity$\uparrow$&validity$\uparrow$&log-LOF$\downarrow$&time$\downarrow$&Acc/F1\\
        \hline
        Wachter&32.94\tiny $\pm$ 0.41&-&0\tiny$\pm$0&0.57\tiny$\pm$0.03&0.01\tiny$\pm$0&99.46/98.56\\
         DiCE&3.57\tiny$\pm$0.04&6.84\tiny$\pm$0.01&1\tiny $\pm$0&0.15\tiny$\pm$0.03&35.93\tiny$\pm$2.34&99.35/98.29\\
         FACE&-&-&-&-&-&-\\
         VCNet&30.03\tiny$\pm$0.09&-&0.93\tiny$\pm$0.01&0.4\tiny$\pm$0.05&3.77e-6\tiny$\pm$3.4e-7&98.44/95.8\\
         CCHVAE&-&-&-&-&-&-\\
         SCD&38.43\tiny$\pm$0.05&8.82\tiny$\pm$0.06&1\tiny$\pm$0&0.39\tiny$\pm$0.03&3.69\tiny$\pm$0.05&99.24/97.17\\
         \textbf{TabDiff}&34.07\tiny$\pm$0.08&7.53\tiny$\pm$0.01&1\tiny$\pm$0&0.19\tiny$\pm$0.002&6.83\tiny$\pm$0.14&99.33/97.21\\
         \textbf{TabDiff-distill.}&30.69\tiny$\pm$0.26&7.71\tiny$\pm$0.05&1\tiny$\pm$0&0.53\tiny$\pm$0.05&0.54\tiny$\pm$0.008&99.28/98.09\\
         \hline
    \end{tabularx}
    \caption{Results for CIC-IDS-2017 dataset, with 1000 attack queries. TabDiff-distill. provides highly valid, and efficient counterfactual explanations. Although maximally sparse results with high validity are provided by DiCE, this behavior was not consistent across other datasets. The value of the sparsity metric of TabDiff models are similar to other datasets when considered as a ratio to the total number of features.}
    \label{tab:cic1}
\end{table}

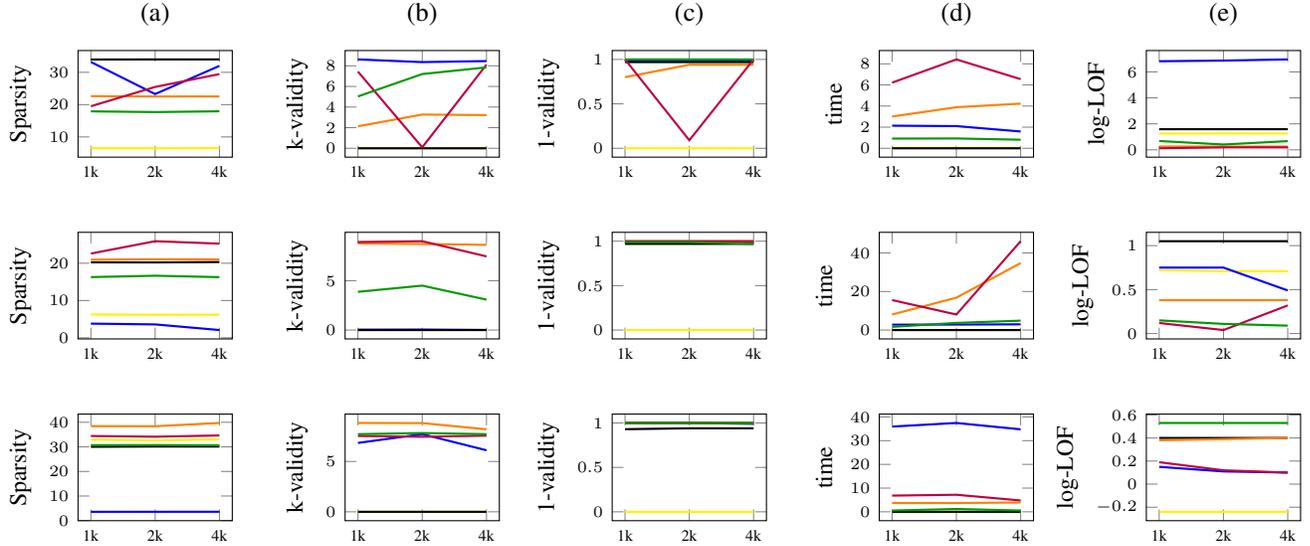
\begin{figure}[t] % Use [t] or [b] or [!ht] depending on placement needs
\centering
\begin{tikzpicture}

\begin{groupplot}[
  group style={
    group size=5 by 3,
     horizontal sep=1.5cm,
        vertical sep=1cm
  },
  width=0.22\textwidth,
  height=3cm,
  xtick={0,1,2},
  xticklabels={1k,2k,4k},
  ylabel={},
  ylabel near ticks,
  tick label style={font=\tiny},
  legend style={at={(0.5,-0.2)}, anchor=north, font=\tiny, legend columns=3},
  cycle list name=mycolorlist,
]

% --- Plot 1 ---
\nextgroupplot[title={(a)},ylabel={\small Sparsity}]
\addplot coordinates {(0,6.51) (1,6.49) (2,6.56)};
\addplot coordinates {(0,33.2) (1,23.27) (2,31.99)};
\addplot coordinates {(0,33.96) (1,33.99) (2,33.99)};
\addplot coordinates {(0,22.63) (1,22.52) (2,22.53)};
\addplot coordinates {(0,19.5) (1,25.47) (2,29.45)};
\addplot coordinates {(0,17.92) (1,17.69) (2,17.94)};

% --- Plot 2 ---
\nextgroupplot[title={(b)},ylabel={\small k-validity}]
\addplot coordinates {(0,0) (1,0) (2,0)};
\addplot coordinates {(0,8.62) (1,8.36) (2,8.46)};
\addplot coordinates {(0,0) (1,0) (2,0)};
\addplot coordinates {(0,2.13) (1,3.27) (2,3.22)};
\addplot coordinates {(0,7.44) (1,0.09) (2,8.12)};
\addplot coordinates {(0,5.03) (1,7.2) (2,7.85)};

\nextgroupplot[title={(c)},ylabel={\small 1-validity}]
\addplot coordinates {(0,0) (1,0) (2,0)};
\addplot coordinates {(0,0.99) (1,0.98) (2,0.99)};
\addplot coordinates {(0,0.97) (1,0.97) (2,0.97)};
\addplot coordinates {(0,0.8) (1,0.94) (2,0.94)};
\addplot coordinates {(0,1) (1,0.09) (2,1)};
\addplot coordinates {(0,1) (1,1) (2,1)};

% time
\nextgroupplot[title={(d)},ylabel={\small time}]
\addplot coordinates {(0,0.04) (1,0.04) (2,0.04)};
\addplot coordinates {(0,2.14) (1,2.1) (2,1.59)};
\addplot coordinates {(0,4e-6) (1,0) (2,0)};
\addplot coordinates {(0,3.01) (1,3.89) (2,4.23)};
\addplot coordinates {(0,6.22) (1,8.42) (2,6.55)};
\addplot coordinates {(0,0.92) (1,0.93) (2,0.81)};

%lof
\nextgroupplot[title={(e)},ylabel={\small log-LOF}]
\addplot coordinates {(0,1.25) (1,1.26) (2,1.26)};
\addplot coordinates {(0,6.82) (1,6.87) (2,6.96)};
\addplot coordinates {(0,1.59) (1,1.59) (2,1.59)};
\addplot coordinates {(0,0.26) (1,0.25) (2,0.25)};
\addplot coordinates {(0,0.12) (1,0.19) (2,0.19)};
\addplot coordinates {(0,0.68) (1,0.41) (2,0.67)};

%=====================================================%

% --- Plot 1 ---
\nextgroupplot[title={},ylabel={\small Sparsity}]
\addplot coordinates {(0,6.25) (1,6.18) (2,6.21)};
\addplot coordinates {(0,3.8) (1,3.59) (2,2.08)};
\addplot coordinates {(0,20.24) (1,20.22) (2,20.32)};
\addplot coordinates {(0,20.95) (1,21.07) (2,21.02)};
\addplot coordinates {(0,22.59) (1,25.91) (2,25.25)};
\addplot coordinates {(0,16.28) (1,16.65) (2,16.27)};

% --- Plot 2 ---
\nextgroupplot[title={},ylabel={\small k-validity}]
\addplot coordinates {(0,0) (1,0) (2,0)};
\addplot coordinates {(0,0.04) (1,0.05) (2,0.01)};
\addplot coordinates {(0,0) (1,0) (2,0)};
\addplot coordinates {(0,8.78) (1,8.72) (2,8.63)};
\addplot coordinates {(0,8.92) (1,9) (2,7.46)};
\addplot coordinates {(0,3.87) (1,4.5) (2,3.08)};

\nextgroupplot[title={},ylabel={\small 1-validity}]
\addplot coordinates {(0,0) (1,0) (2,0)};
\addplot coordinates {(0,0.99) (1,0.98) (2,0.99)};
\addplot coordinates {(0,0.97) (1,0.97) (2,0.97)};
\addplot coordinates {(0,1) (1,1) (2,1)};
\addplot coordinates {(0,1) (1,1) (2,1)};
\addplot coordinates {(0,0.99) (1,0.99) (2,0.97)};

% time
\nextgroupplot[title={},ylabel={\small time}]
\addplot coordinates {(0,0.007) (1,0.008) (2,0.008)};
\addplot coordinates {(0,2.8) (1,2.84) (2,2.97)};
\addplot coordinates {(0,3.7e-6) (1,2e-6) (2,1e-6)};
\addplot coordinates {(0,8.06) (1,16.81) (2,34.75)};
\addplot coordinates {(0,15.55) (1,8.05) (2,46.1)};
\addplot coordinates {(0,1.65) (1,3.7) (2,4.87)};

%lof
\nextgroupplot[title={},ylabel={\small log-LOF}]
\addplot coordinates {(0,0.72) (1,0.71) (2,0.71)};
\addplot coordinates {(0,0.75) (1,0.75) (2,0.49)};
\addplot coordinates {(0,1.05) (1,1.05) (2,1.05)};
\addplot coordinates {(0,0.38) (1,0.38) (2,0.38)};
\addplot coordinates {(0,0.12) (1,0.04) (2,0.32)};
\addplot coordinates {(0,0.15) (1,0.11) (2,0.09)};

%=======================================================%
% --- Plot 1 ---
\nextgroupplot[title={},ylabel={\small Sparsity}]
\addplot coordinates {(0,32.94) (1,32.62) (2,33.06)};
\addplot coordinates {(0,3.57) (1,3.58) (2,3.59)};
\addplot coordinates {(0,30.03) (1,30.14) (2,30.13)};
\addplot coordinates {(0,38.43) (1,38.39) (2,39.72)};
\addplot coordinates {(0,34.42) (1,34.13) (2,34.64)};
\addplot coordinates {(0,30.69) (1,30.72) (2,30.61)};

% --- Plot 2 ---
\nextgroupplot[title={},ylabel={\small k-validity}]
\addplot coordinates {(0,0) (1,0) (2,0)};
\addplot coordinates {(0,6.84) (1,7.71) (2,6.11)};
\addplot coordinates {(0,0) (1,0) (2,0)};
\addplot coordinates {(0,8.82) (1,8.81) (2,8.19)};
\addplot coordinates {(0,7.53) (1,7.45) (2,7.56)};
\addplot coordinates {(0,7.71) (1,7.82) (2,7.71)};

\nextgroupplot[title={},ylabel={\small 1-validity}]
\addplot coordinates {(0,0) (1,0) (2,0)};
\addplot coordinates {(0,1) (1,1) (2,0.99)};
\addplot coordinates {(0,0.93) (1,0.94) (2,0.94)};
\addplot coordinates {(0,1) (1,1) (2,1)};
\addplot coordinates {(0,1) (1,1) (2,1)};
\addplot coordinates {(0,1) (1,1) (2,1)};

% time
\nextgroupplot[title={},ylabel={\small time}]
\addplot coordinates {(0,0.01) (1,0.01) (2,0.01)};
\addplot coordinates {(0,35.93) (1,37.45) (2,34.77)};
\addplot coordinates {(0,4.58e-6) (1,1.02e-6) (2,4.22e-6)};
\addplot coordinates {(0,3.69) (1,3.71) (2,3.92)};
\addplot coordinates {(0,6.83) (1,7.19) (2,4.78)};
\addplot coordinates {(0,0.54) (1,1.19) (2,0.54)};

%lof
\nextgroupplot[title={},ylabel={\small log-LOF}]
\addplot coordinates {(0,-0.24) (1,-0.24) (2,-0.24)};
\addplot coordinates {(0,0.15) (1,0.11) (2,0.1)};
\addplot coordinates {(0,0.4) (1,0.4) (2,0.4)};
\addplot coordinates {(0,0.38) (1,0.39) (2,0.4)};
\addplot coordinates {(0,0.19) (1,0.12) (2,0.10)};
\addplot coordinates {(0,0.53) (1,0.53) (2,0.53)};

\end{groupplot}
\end{tikzpicture}
\caption{Comparison of the evaluated metrics of the counterfactual explanations with increasing attack pool sizes. Each row depicts the results for UNSW-NB15, CIC-DDoS-2019 and CIC-IDS-2017, respectively. The colors of the line plots map as following; Wachter (yellow), DiCE (blue), VCNet (black), SCD (orange), TabDiff (purple), and TabDiff-distill (green). Simple methods such as Wachter remain non-performative, while generative methods, specifically the proposed method (TabDiff-distill.) shows consistent performance in terms of validity, plausibility, and efficiency.} 
\label{fig1}
\end{figure}

The line plots in figure~\ref{fig1} depict the performance of the same metrics on increasing sizes of attack pools. Table~\ref{tab:real} show a few samples of counterfactual explanations generated for the CIC-DDoS-2019 dataset.

\pgfplotscreateplotcyclelist{mycolorlist}{
  {yellow, thick},
  {blue, thick},
  {black, thick},
  {orange, thick},
  {purple, thick},
  {green!60!black, thick},
}

\subsection{Global counterfactual rules}
The above results present a holistic view of the generated counterfactual explanations across different methods, in terms of quantitative metrics. We then look at how the generated counterfactual explanations can be utilized for identifying and drawing countermeasures against attacks. A trivial approach would be to compare the feature values of the queries and their respective counterfactual explanations. As an alternative approach, we propose a simple counterfactual rule approach. First we extract separate a set of specific attack data points from the train set and create a new version of the original training set without the specific attack . This step ensures that the specific attack was not seen during the model training phase. These extracted attack data points can be regarded as an unknown (or 'zero-day') attack. As example use cases, we consider for two attacks; the first one is 'Analysis' attack from the UNSW-NB15 dataset, and the second one 'LDAP DDoS' attack from the CIC-DDoS-2019 dataset (these two attacks were picked based on the performance of the classifier model of them). We first evaluate the performance of the black-box classifier on the unseen attack to confirm that it is at an acceptable level (for the 'Analysis' attack we obtain an Accuracy of 81.81\%). Counterfactual explanations are then generated for this specific attack's test set using the proposed diffusion method. Finally, global counterfactual rules are extracted using a simple decision-tree (the algorithm for extracting the counterfactual rules are presented as a pseudo code in algorithm~\ref{alg:rule} in appendix~\ref{app2}). A few such extracted rules for the 'Analysis' attack are presented below. 

\begin{itemize}
    \item 'state$<$2', 'proto$<=$46', 'djit$<=$91402.99', 'proto$>$0’
    \item 'state$>$1','trans\_depth$<=$1.0','dload$<=$2579956.5',
    'state$<$7’, proto$>$25','proto$<$27’
\end{itemize}

Each rule provides bounds that filters benign data from the attack data based on a set of important features. The feature descriptions are as below~\cite{moustafa_unsw-nb15_2015}.
Moreover, we compare the SHAP plot obtained for the same attack and the corresponding counterfactual data points (Figure~\ref{fig3}(a)). It is observed that the derived global rules can provide value bounds that can filter benign data from attack data while also being consistent with the important features identified by the SHAP explanations.\\
We further compare these rules against rules obtained by replacing the counterfactual explanations with randomly sampled benign data from the training dataset (Figure~\ref{fig3}(b)). It is seen that the rules presented by the training data can be different in terms of the features compared to the rules derived from using the counterfactual explanations. In other words, rules derived using training data pick common features across different attacks such as 'response\_body\_len' and 'is\_ftp\_login' which can be relevant to different types of attacks.

\begin{itemize}
    \item  dur$<=$ 15.34,sload $<=$ 72363628.0,response\_body\_len $<=$ 141.5,ct\_flw\_http\_mthd $<=$ 2.5,dtcpb $<=$ 4201986176.0,ct\_dst\_sport\_ltm $<=$ 3.5,is\_ftp\_login $<=$ 2.5
\end{itemize}

The difference between the way of deriving rules (i.e.- using counterfactuals vs training data) is more pronounced for the 'LDAP DDoS' attack. The SHAP plot and the corresponding rules show that the features identified using the counterfactual explanations is more informative than the rules derived using training data (Figure~\ref{fig4}(a),(b)). The rules extracted from the counterfactual explanations followed by the rules extracted from training data are shown below.
With counterfactual explanations,
\begin{itemize}
    \item 'Fwd Header Length'$<=$205.00,  'ACK Flag Count'$>$0, 'Protocol'$<$=11,  'CWE Flag Count'$<$1,  'Protocol'$>$3,  'Idle Std'$<=$31137002.0\\
  \item 'Fwd Header Length'$<=$205.00,  'ACK Flag Count'$>$0.5,  'Protocol'$<=$11,  'CWE Flag Count'$>$0,  'RST Flag Count'$<$1,  'Fwd PSH Flags'$>$0,  'SYN Flag Count'$>$0'
\end{itemize}

With training data (only a single rule is available),
\begin{itemize}
    \item 'Flow IAT Mean'$>$0.625, 'Fwd Packet Length Min'$<=$1148.498'
\end{itemize}

It is observed that the rules extracted from the training data contains features such as 'Flow IAT Mean', and 'Fwd Packet Length Min', which are general to all types of DDoS attacks~\cite{sharafaldin_developing_2019}, whereas rules extracted from counterfactuals contain protocol specific rules. As seen in table~\ref{tab:real}, the diversity of the generated counterfactual explanations provides different explanations that act as different viewpoints. For example, changing 'CWE Flag Count' to 1 or keeping it 0, and increasing 'Fwd Packet Length Std' or keeping it the same are two options when generating counterfactual explanations (in row 2 and 4). In addition, global rules can be generated even with a small number of attack queries since diversity of counterfactual explanations provide multiple benign samples per each attack query.

\pgfplotsset{
    var1/.style={mark=diamond*, only marks},
    var2/.style={mark=triangle*, only marks},
    var3/.style={mark=Mercedes star, only marks},
    var4/.style={mark=star, only marks},
    var5/.style={mark=*, only marks},
    var6/.style={mark=square*, only marks},
}

% Define consistent colors for 3 settings
\definecolor{settingA}{RGB}{31,119,180}  % blue
\definecolor{settingB}{RGB}{255,127,14}  % orange
\definecolor{settingC}{RGB}{44,160,44}   % green

\begin{figure}[H]
\centering

% === Metric 1 ===
\begin{minipage}{\textwidth}
\centering
\begin{tikzpicture}
\begin{axis}[
    title=(a),
    xlabel={sparsity},
    ylabel={time},
    xticklabel style={rotate=45, font=\tiny},
    yticklabel style={font=\tiny},
    width=0.49\textwidth,
    height=4cm,
    grid=both,
]
\addplot+[var1, color=settingA] coordinates {(6.51,0.04)};
\addplot+[var1, color=settingB] coordinates {(6.25,7.8e-3)};
\addplot+[var1, color=settingC] coordinates {(32.94,0.01)};
\addplot+[var2, color=settingA] coordinates {(33.2,2.14)};
\addplot+[var2, color=settingB] coordinates {(3.8,2.8)};
\addplot+[var2, color=settingC] coordinates {(3.75,35.93)};
\addplot+[var3, color=settingA] coordinates {(33.96,4.35e-6)};
\addplot+[var3, color=settingB] coordinates {(20.24,4.58e-6)};
\addplot+[var3, color=settingC] coordinates {(30.03,3.77e-6)};
\addplot+[var4, color=settingA] coordinates {(22.63,3.01)};
\addplot+[var4, color=settingB] coordinates {(20.95,8.06)};
\addplot+[var4, color=settingC] coordinates {(38.43,3.69)};
\addplot+[var5, color=settingA] coordinates {(19.5,6.22)};
\addplot+[var5, color=settingB] coordinates {(22.59,15.55)};
\addplot+[var5, color=settingC] coordinates {(34.42,6.83)};
\addplot+[var6, color=settingA] coordinates {(17.92,0.92)};
\addplot+[var6, color=settingB] coordinates {(16.28,1.65)};
\addplot+[var6, color=settingC] coordinates {(30.69,0.54)};
\pgfplotsextra{
    \pgfmathsetmacro{\xmid}{(\pgfkeysvalueof{/pgfplots/xmin} + \pgfkeysvalueof{/pgfplots/xmax}) / 2}
    \pgfmathsetmacro{\ymid}{(\pgfkeysvalueof{/pgfplots/ymin} + \pgfkeysvalueof{/pgfplots/ymax}) / 2}

    \begin{scope}
        \draw[red,thick,dashed]
            (axis cs:\pgfkeysvalueof{/pgfplots/xmin}, \pgfkeysvalueof{/pgfplots/ymin})
            rectangle
            (axis cs:\xmid, \ymid);
    \end{scope}
}
\end{axis}
\end{tikzpicture}
\end{minipage}
%
% === Metric 2 ===
\begin{minipage}{\textwidth}
\centering
\begin{tikzpicture}
\begin{axis}[
    title=(b),
    xticklabel style={rotate=45, font=\tiny},
    yticklabel style={font=\tiny},
    xlabel={validity},
    ylabel={time},
    width=0.49\textwidth,
    height=4cm,
    grid=both,
]
\addplot+[var1, color=settingA] coordinates {(0,0.04)};
\addplot+[var1, color=settingB] coordinates {(0,7.8e-3)};
\addplot+[var1, color=settingC] coordinates {(0,0.01)};
\addplot+[var2, color=settingA] coordinates {(0.99,2.14)};
\addplot+[var2, color=settingB] coordinates {(0.04,2.8)};
\addplot+[var2, color=settingC] coordinates {(1,35.93)};
\addplot+[var3, color=settingA] coordinates {(0.97,4.35e-6)};
\addplot+[var3, color=settingB] coordinates {(0.97,4.58e-6)};
\addplot+[var3, color=settingC] coordinates {(0.93,3.77e-6)};
\addplot+[var4, color=settingA] coordinates {(0.8,3.01)};
\addplot+[var4, color=settingB] coordinates {(1,8.06)};
\addplot+[var4, color=settingC] coordinates {(1,3.69)};
\addplot+[var5, color=settingA] coordinates {(1,6.22)};
\addplot+[var5, color=settingB] coordinates {(1,15.55)};
\addplot+[var5, color=settingC] coordinates {(1,6.83)};
\addplot+[var6, color=settingA] coordinates {(1,0.92)};
\addplot+[var6, color=settingB] coordinates {(0.99,1.65)};
\addplot+[var6, color=settingC] coordinates {(1,0.54)};
% repeat for var2 to var6...
\pgfplotsextra{
    \pgfmathsetmacro{\xmid}{(\pgfkeysvalueof{/pgfplots/xmin} + \pgfkeysvalueof{/pgfplots/xmax}) / 2}
    \pgfmathsetmacro{\ymid}{(\pgfkeysvalueof{/pgfplots/ymin} + \pgfkeysvalueof{/pgfplots/ymax}) / 2}

    \begin{scope}
        \draw[red,thick,dashed]
            (axis cs:\xmid, \pgfkeysvalueof{/pgfplots/ymin})
            rectangle
            (axis cs:\pgfkeysvalueof{/pgfplots/xmax}, \ymid);
    \end{scope}
}
\end{axis}
\end{tikzpicture}
\end{minipage}

%
% % === Metric 3 ===
\begin{minipage}{\textwidth}
\centering
\begin{tikzpicture}
\begin{axis}[
    title=(c),
    xticklabel style={rotate=45, font=\tiny},
    yticklabel style={font=\tiny},
    xlabel={log-LOF},
    ylabel={time},
    width=0.49\textwidth,
    height=4cm,
    grid=both,
]
\addplot+[var1, color=settingA] coordinates {(1.25,0.04)};
\addplot+[var1, color=settingB] coordinates {(0.72,7.8e-3)};
\addplot+[var1, color=settingC] coordinates {(0.57,0.01)};
\addplot+[var2, color=settingA] coordinates {(6.82,2.14)};
\addplot+[var2, color=settingB] coordinates {(0.72,2.8)};
\addplot+[var2, color=settingC] coordinates {(0.15,35.93)};
\addplot+[var3, color=settingA] coordinates {(1.59,4.35e-6)};
\addplot+[var3, color=settingB] coordinates {(1.05,4.58e-6)};
\addplot+[var3, color=settingC] coordinates {(0.4,3.77e-6)};
\addplot+[var4, color=settingA] coordinates {(0.26,3.01)};
\addplot+[var4, color=settingB] coordinates {(0.38,8.06)};
\addplot+[var4, color=settingC] coordinates {(0.39,3.69)};
\addplot+[var5, color=settingA] coordinates {(0.12,6.22)};
\addplot+[var5, color=settingB] coordinates {(0.12,15.55)};
\addplot+[var5, color=settingC] coordinates {(1.19,6.83)};
\addplot+[var6, color=settingA] coordinates {(0.68,0.92)};
\addplot+[var6, color=settingB] coordinates {(0.15,1.65)};
\addplot+[var6, color=settingC] coordinates {(0.53,0.54)};
% repeat for var2 to var6...
\pgfplotsextra{
    \pgfmathsetmacro{\xmid}{(\pgfkeysvalueof{/pgfplots/xmin} + \pgfkeysvalueof{/pgfplots/xmax}) / 2}
    \pgfmathsetmacro{\ymid}{(\pgfkeysvalueof{/pgfplots/ymin} + \pgfkeysvalueof{/pgfplots/ymax}) / 2}

    \begin{scope}
        \draw[red,thick,dashed]
            (axis cs:\pgfkeysvalueof{/pgfplots/xmin}, \pgfkeysvalueof{/pgfplots/ymin})
            rectangle
            (axis cs:\xmid, \ymid);
    \end{scope}
}
\end{axis}
\end{tikzpicture}
\end{minipage}
% \vspace{-1em} 
% LEGEND
\begin{minipage}{\textwidth}
\centering
\begin{tikzpicture}
\begin{axis}[
    hide axis,
    xmin=0, xmax=1,
    ymin=0, ymax=1,
    legend style={
        font=\footnotesize,
        draw=none,
        at={(0.5,1.1)},
        anchor=south,
        legend columns=3,
        column sep=1em,
    }
]
\addlegendimage{var1, black}
\addlegendentry{Wachter}
\addlegendimage{var2, black}
\addlegendentry{DiCE}
\addlegendimage{var3, black}
\addlegendentry{VCNet}
\addlegendimage{var4, black}
\addlegendentry{SCD}
\addlegendimage{var5, black}
\addlegendentry{TabDiff}
\addlegendimage{var6, black}
\addlegendentry{TabDiff-distill.}
\addlegendimage{only marks, color=settingA, mark=*}
\addlegendentry{UNSW-NB15}
\addlegendimage{only marks, color=settingB, mark=*}
\addlegendentry{CIC-DDoS-2019}
\addlegendimage{only marks, color=settingC, mark=*}
\addlegendentry{CIC-IDS-2017}
\end{axis}
\end{tikzpicture}    
\end{minipage}
\vspace{-15em}
\caption{Scatter plots representing the behavior of sparsity (a), validity (b) and log-LOF(c) against time (from tables~\ref{tab:unsw1},\ref{tab:ddos1},\ref{tab:cic1}). For each plot, points in (a) - low left corner, (b) - low right corner, (c) - low left corner are the preferred in terms of performance (marked with a dashed, red rectangle in each plot). Proposed methods, especially TabDiff-distill falls close to these regions. Shapes = Counterfactual methods; Colors = Datasets}
\label{scatter}
\end{figure}
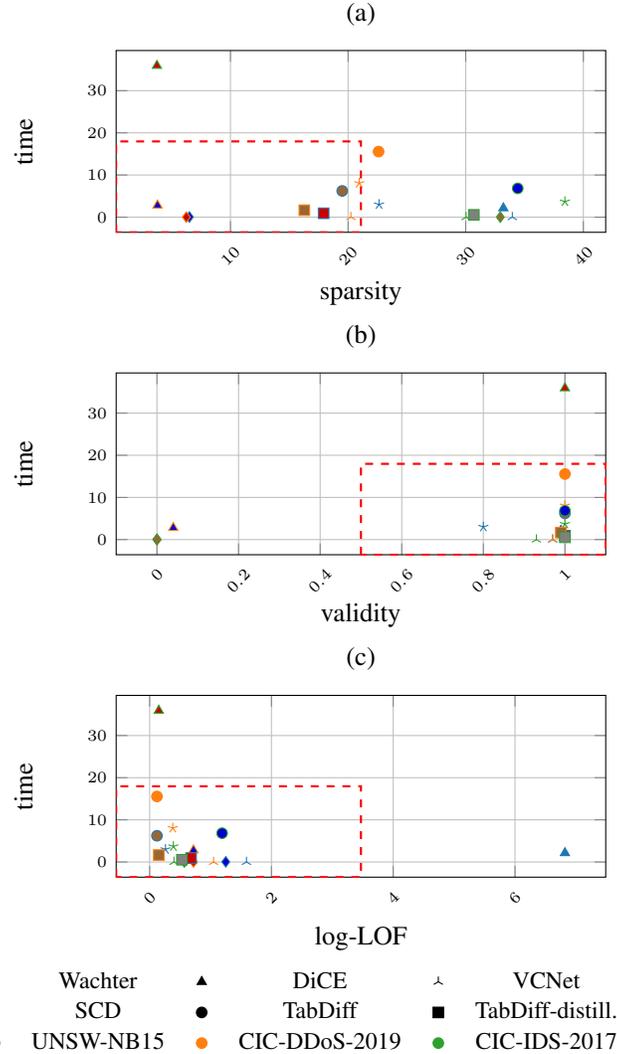

\section{Discussion}
The results reveal that the proposed diffusion based methods, specifically the TabDiff-distill. version offer the best counterfactual explanations in-terms of sparsity, plausibility, validity and efficiency for UNSW-NB15 and CIC-DDoS-2019 datasets. It can be also observed that TabDiff-distill. trades-off the improvements of sparsity, and efficiency from the vanilla TabDiff model with the anomaly score. This behavior can be attributed to the trade-off of the generation quality for the efficiency (by reducing the number of diffusion steps) of progressive distillation~\cite{salimans_progressive_2022}. DiCE method provides the best explanation in term of sparsity for the CIC-IDS-2017 with good plausibility. This behavior contradicts its behavior for the other two datasets which returns explanations with a high anomaly score (high log-LOF). DiCE returning anomalous explanations can be expected due to the nature of the algorithm - which utilizes random perturbations. The behavior for CIC-IDS-2017 could be attributed to that DiCE manages to generate counterfactuals which are close to the data manifold and with minimal modifications through random perturbations. However, this behavior is not consistent across all the datasets.\\

Although the performance is not consistent, VCNet provides plausible and valid counterfactual explanations for two datasets. The inconsistent behavior can be attributed to the lack of an explicit optimization method available for VCNet (i.e.- VCNet resorts to conditional sampling from the CVAE model, but do not perform any optimization in the latent space). Usually, generative methods such as VCNet, and TabDiff tend to generate explanations in the form of real data samples which is evident from their plausibility and the nature of their sparsity, while random perturbation methods aim at generating explanations with minimal modifications to the original. These random perturbations may not be consistently successfull at all instances, as it was observed in UNSW-NB15 and CIC-DDoS-2019 datasets. On the other hand, while simple counterfactual algorithms such as  Wachter is unsuccessful at producing counterfactual explanations, more advanced methods such as FACE and CCHVAE also fail to do so. This raises the concern that not all available counterfactual explanation methods that are primarily focused on algorithmic recourse, would be suitable for a complex domain such as NIDS.\\ 

The global counterfactual rules derived from counterfactual explanations provide insights into the nature of the 'Analysis' attack. Features such as 'djit' (destination jitter), 'dload' (destination load) and 'trans\_depth' (transaction depth) can be abnormal for the 'Analysis' attack, as an attacker sends probing packets to target devices with high variations and abrupt connections (e.g.- prematurely ending an http connection). Moreover, the proposed technique of obtaining such rules using counterfactual explanations eliminate the need of accessing the training data to derive explanations and rules. In addition, using training data to derive such rules can present more generic rules, whereas the counterfactuals based rules can be more specific. This can be observed from the features present in the training data based rules, which contains 'response\_body\_len'(length of the reponse body) which can be more common to different types of attacks. Another example is 'is\_ftp\_login'(FTP login attempt) which can be common to other attack types. However, one shortcoming of this rule derivation is that the supervised approach requires a acceptable classifier performance of the black-box classifier on unseen attacks in order to derive valid explanations and the rules. A straightforward utility of the derived rules is providing actionable explanations that can act as countermeasures against incoming attacks by filtering out packets based on the identified features and their value bounds. However, we do not evaluate the efficacy of the filtering capability of the rules as it is out of scope of this work.

% \begin{figure}
%     \centering
%     \includegraphics[scale=0.4]{shap.png}
%     \caption{SHAP feature importance plot obtained using generated counterfactual explanations for the 'Analysis' attack}
%     \label{fig2}
% \end{figure}
% \begin{figure}
%     \centering
%     \includegraphics[scale=0.2]{shap2.png}
%     \caption{SHAP feature importance plot obtained using training data for the 'Analysis' attack}
%     \label{fig3}
% \end{figure}

% \begin{figure}[htbp]
%   \centering
%   \begin{subfigure}[t]{0.49\columnwidth}
%     \centering
%     \includegraphics[width=\linewidth]{shap.png}
%     \caption{}
%   \end{subfigure}
%   \hfill
%   \begin{subfigure}[t]{0.49\columnwidth}
%     \centering
%     \includegraphics[width=\linewidth]{shap2.png}
%     \caption{}
%   \end{subfigure}
%   \caption{SHAP feature importance plot obtained using generated  (a) - counterfactual explanations, and  (b) - using training data for the 'Analysis' attack. The SHAP values shows a ranking of how each feature contributes towards the average model prediction, across the dataset. For example, there are certain feature values in the 'state' feature, that negatively contributes to the model prediction (blue), and positively contributes (red) to the average model prediction.}
%   \label{fig3}
% \end{figure}

\begin{figure}[H]
  \centering
  \begin{minipage}[t]{0.48\linewidth}
    \centering
    \includegraphics[width=0.7\linewidth,height=6cm]{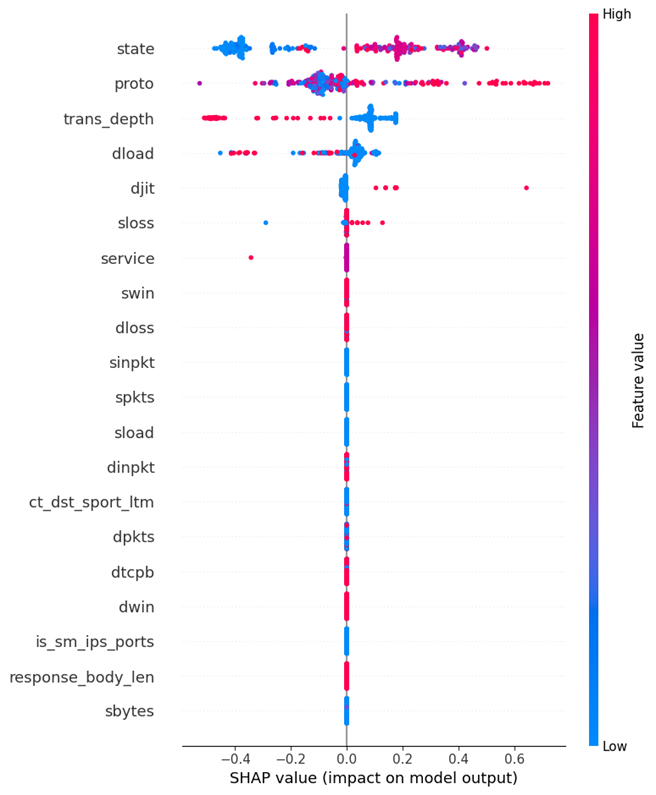}
    \caption*{(a)}
  \end{minipage}
  \hfill
  \begin{minipage}[t]{0.48\linewidth}
    \centering
    \includegraphics[width=0.7\linewidth,height=6cm]{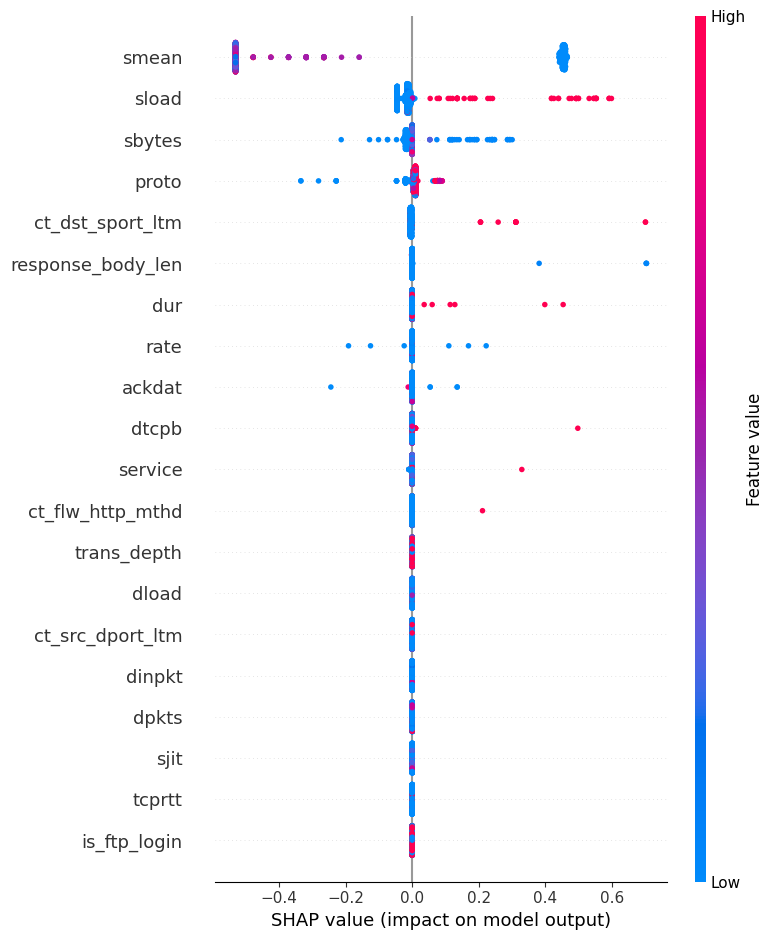}
    \caption*{(b)}
  \end{minipage}
  \caption{SHAP feature importance plot obtained using generated  (a) - counterfactual explanations, and  (b) - using training data for the 'Analysis' attack. The SHAP values shows a ranking of how each feature contributes towards the average model prediction, across the dataset. For example, there are certain feature values in the 'state' feature, that negatively contributes to the model prediction (blue), and positively contributes (red) to the average model prediction.}
  \label{fig3}
\end{figure}

% \begin{figure}
%     \centering
%     \includegraphics[scale=0.2]{shap3.png}
%     \caption{SHAP feature importance plot obtained using generated counterfactuals for the 'DDoS LDAP' attack}
%     \label{fig4}
% \end{figure}

% \begin{figure}
%     \centering
%     \includegraphics[scale=0.2]{shap4.png}
%     \caption{SHAP feature importance plot obtained using training data for the 'DDoS LDAP' attack}
%     \label{fig5}
% \end{figure}

% \begin{figure}[htbp]
%   \centering
%   \begin{subfigure}
%     \centering
%     \includegraphics[width=0.49\linewidth]{shap3.png}
%     \caption{}
%   \end{subfigure}
%   \hfill
%   \begin{subfigure}
%     \centering
%     \includegraphics[width=0.49\linewidth]{shap4.png}
%     \caption{}
%   \end{subfigure}
%   \caption{SHAP feature importance plot obtained (a) - using generated counterfactuals, and (b) - using training data for the 'DDoS LDAP' attack}
%   \label{fig4}
% \end{figure}

\begin{figure}[H]
  \centering
  \begin{minipage}[t]{0.48\linewidth}
    \centering
    \includegraphics[width=0.7\linewidth,height=6cm]{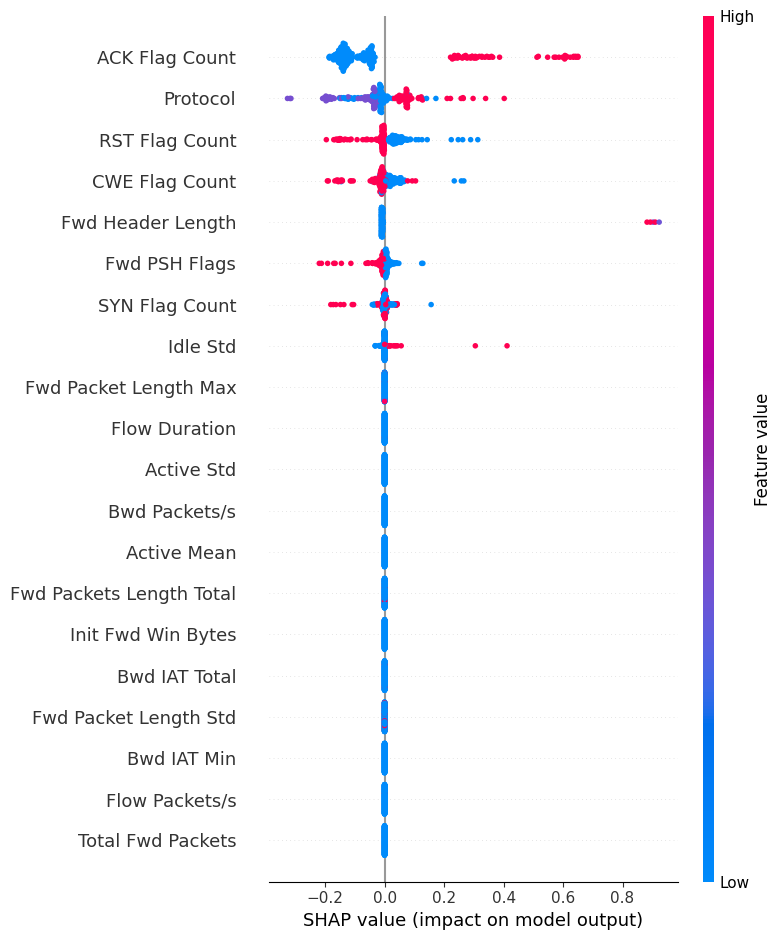}
    \caption*{(a)}
  \end{minipage}
  \hfill
  \begin{minipage}[t]{0.48\linewidth}
    \centering
    \includegraphics[width=0.7\linewidth,height=6cm]{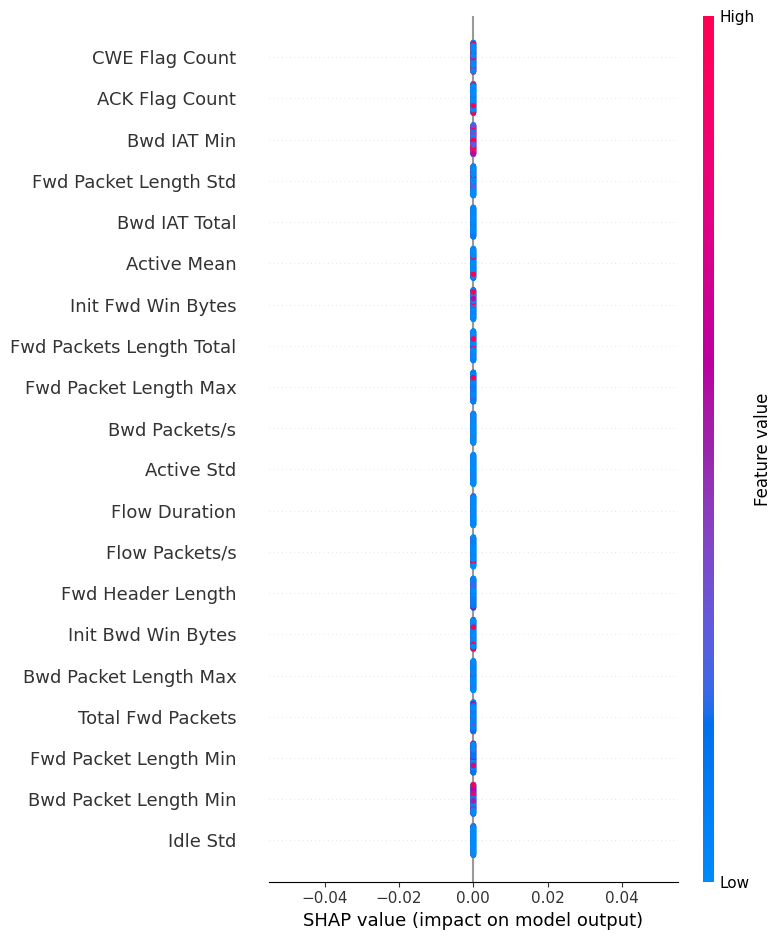}
    \caption*{(b)}
  \end{minipage}
  \caption{SHAP feature importance plot obtained (a) - using generated counterfactuals, and (b) - using training data for the 'DDoS LDAP' attack}
  \label{fig4}
\end{figure}

\section{Conclusion}
We carried out an empirical evaluation of several existing counterfactual explanation algorithms for NIDS, and proposed novel, diffusion based counterfactual generation approaches. The proposed methods can efficiently generate valid, diverse and plausible counterfactual explanations for network intrusion datasets. In addition, we presented a simple approach to obtain global counterfactual rules that can potentially act as countermeasures against incoming intrusion attacks.We anticipate that this work will stimulate further discourse on the application of counterfactual explanations within the domain of NIDS, which is majorly overlooked at present. Evaluating the efficacy of derived global rules in different practical scenarios is left as future work. In addition, the proposed methods can be extended to other NID datasets in the future. 
\appendix

\section{Theoretical details}
\label{app1}

% \subsection{Denoising Diffusion Probabilistic Models for Tabular data}

\subsection{Progressive Distillation of DDPM}

We adopt the progressive distillation proposed by Salimans et al.~\cite{salimans_progressive_2022}. As per their findings, we reformulate the conventional diffusion processes in DDPM with 'v-diffusion' first. v-diffusion is a angular reparameterization of the original diffusion process, and can be represented mathematically as in Eq.~\ref{eq:vdiff}.

\begin{equation}
\label{eq:vdiff}
\mathbf{v}_\phi \equiv \frac{d\mathbf{z}_\phi}{d\phi} 
= \frac{d\cos(\phi)}{d\phi} \mathbf{x} 
+ \frac{d\sin(\phi)}{d\phi} \boldsymbol{\epsilon} 
= \cos(\phi) \boldsymbol{\epsilon} - \sin(\phi) \mathbf{x}.
\end{equation}

where,

\begin{equation}
\phi_t = \arctan\left( \frac{\sqrt{(1-\bar\alpha_t)}}{\bar\alpha_t} \right)
\end{equation}

Then, the goal of the diffusion model becomes approximating the v-value (Eq.~\ref{eq:vdiff2}). 

\begin{equation}
\label{eq:vdiff2}
\hat{\mathbf{v}}_\theta(\mathbf{z}_\phi) \equiv 
\cos(\phi)\, \hat{\boldsymbol{\epsilon}}_\theta(\mathbf{z}_\phi) 
- \sin(\phi)\, \hat{\mathbf{x}}_\theta(\mathbf{z}_\phi),
\end{equation}

This reformulation is carried out as it has been shown that progressive distillation technique works better with v-diffusion~\cite{salimans_progressive_2022}. The distillation starts with the original diffusion model (with 2500 steps), and iteratively tries approximating consecutive two diffusion steps with 1 step. This process is carried out until the student model (distilled model) can approximate 2500 steps with 250 steps.

\section{Details of the classifier models and other hyperparameters}
\label{app2}

\subsection{Black box classifier}

The black-box classifier (feed-forward neural network) contains the below architecture and hyperparameter choices except for the VCNet model. 

\begin{table}[H]
    \centering
    \begin{tabular}{c|c}
        \hline
         Parameter&Value  \\
         \hline
         No. of Hidden Units&128,64,32\\
         Optimizer and lr.&Adam, 5e-4\\
         Epochs&600, (300 - CICIDS-2017)\\
         \hline
    \end{tabular}
    \caption{Architecture and the hyperparameters used for the black-box classifier models}
    \label{tab:ffn1}
\end{table}

\begin{table*}[ht]
\centering
\scriptsize
\caption{Combined Features and Predictions for Generated Counterfactual Explanations}
\begin{tabular}{l|rrrrrr}
\toprule
Feature & orig & 1 & 2 & 3 & 4 & 5 \\
\midrule
Init Bwd Win Bytes         & -1.000000 & -1.000000 & -1.000000 & -1.000000 & -1.000000 & -1.000000 \\
Bwd Packet Len Max         & 0.000000  & 0.000000  & 0.000000  & 0.000000  & 0.000000  & 0.000000  \\
Fwd Act Data Packets       & 1.000000  & 0.000000  & 0.000000  & 0.000000  & 0.000000  & 0.000000  \\
Bwd Header Len             & 0.000000  & 0.000000  & 0.000000  & 0.000000  & 0.000000  & 0.000000  \\
Bwd IAT Min                & 0.000000  & 0.000000  & 0.000000  & 0.000000  & 0.000000  & 0.000000  \\
Flow IAT Mean              & 3.000004  & 0.500000  & 0.500000  & 0.500000  & 0.500000  & 0.500000  \\
Flow IAT Min               & 3.000000  & 0.000000  & 0.000000  & 0.000000  & 0.000000  & 0.000000  \\
Total Fwd Packets          & 2.000000  & 1.000000  & 1.000000  & 1.000000  & 1.000000  & 1.000000  \\
Bwd IAT Total              & 0.000000  & 0.000000  & 0.000000  & 0.000000  & 0.000000  & 0.000000  \\
Total Bwd Packets          & 0.000000  & 0.000000  & 0.000000  & 0.000000  & 0.000000  & 0.000000  \\
Fwd Packet Len Min         & 580.999972 & 0.000000 & 0.000000 & 0.000000 & 0.000000 & 0.000000 \\
Fwd Header Len             & 64.000000  & 0.000000 & 0.000000 & 0.000000 & 0.000000 & 0.000000 \\
Down/Up Ratio              & 0.000000  & 0.000000  & 0.000000  & 0.000000  & 0.000000  & 0.000000  \\
Idle Std                   & 0.000000  & 0.000000  & 0.000000  & 0.000000  & 0.000000  & 0.000000  \\
Flow Packets/s             & 666667.560736 & 0.042459 & 0.042459 & 0.042459 & 0.042459 & 0.042459 \\
Fwd Packet Length Max      & 581.000056 & 0.000000 & 0.000000 & 0.000000 & 0.000000 & 0.000000 \\
Flow Duration              & 3.000000  & 1.000000 & 1.000000 & 1.000000 & 1.000000 & 1.000000 \\
Init Fwd Win Bytes         & -1.000000 & -1.000000 & -1.000000 & 65535.000000 & -1.000000 & -1.000000 \\
Active Std                 & 0.000000  & 0.000000  & 0.000000  & 0.000000  & 0.000000  & 0.000000  \\
Fwd Packet Length Total    & 1162.000245 & 0.000000 & 0.000000 & 77926.000000 & 77926.000000 & 0.000000 \\
Active Mean                & 0.000000  & 0.000000  & 0.000000  & 0.000000  & 0.000000  & 0.000000  \\
Bwd Packets/s              & 0.000000  & 0.000000  & 0.000000  & 0.000000  & 0.000000  & 0.000000  \\
Flow Bytes/s               & 387333669.424055 & 0.000000 & 0.000000 & 0.000000 & 0.000000 & 0.000000 \\
Fwd Packet Length Std      & 0.000000  & 0.000000  & 0.000000  & 1150.217529 & 1150.217529 & 0.000000 \\
Bwd IAT Mean               & 0.000000  & 0.000000  & 0.000000  & 0.000000  & 0.000000  & 0.000000  \\
Protocol                   & 17.000000 & 0.000000  & 0.000000  & 17.000000 & 0.000000  & 0.000000  \\
Fwd PSH Flags              & 0.000000  & 0.000000  & 0.000000  & 1.000000  & 1.000000  & 6.000000  \\
SYN Flag Count             & 0.000000  & 0.000000  & 1.000000  & 0.000000  & 1.000000  & 1.000000  \\
RST Flag Count             & 0.000000  & 0.000000  & 0.000000  & 1.000000  & 0.000000  & 1.000000  \\
ACK Flag Count             & 0.000000  & 1.000000  & 0.000000  & 1.000000  & 0.000000  & 0.000000  \\
CWE Flag Count             & 0.000000  & 1.000000  & 0.000000  & 1.000000  & 1.000000  & 0.000000  \\
Prediction                 & 1         & 1         & 0         & 1         & 0         & 0         \\
\bottomrule
\end{tabular}
\label{tab:real}
\end{table*}

\subsection{Rules extraction}

The global rules extraction is carried out as below. A simple decision tree is used to the benign data points (counterfactual explanations generated) and their respective attack queries as inputs. The decision tree is created with default parameters and with no restrictions on the tree depth.

\begin{algorithm}
\caption{Simple Rule Extraction}\label{alg:rule}
\begin{algorithmic}
\Function{rule extraction}{$x_{CF}, x$}
    \State Encode categorical features as labels in-order to comply with the decision tree model\\
    \State Train a decision tree classifier ($F_{DT}$) with $x_{CF}$ and their original attack queries.\\
    \State Get nodes with high purity for benign data points (i.e.- >0.9) ($n_{p>0.9}$).\\
    \State Get tree paths that include $n_{p>0.9}$ and reach the leaves that have benign classification.\\
    \State Remove redundant rules by parsing the tree paths.
\EndFunction
\end{algorithmic}
\end{algorithm}
%Bibliography
\bibliographystyle{unsrt}  
\bibliography{paper.bib}

\end{document}